\documentclass[11pt]{amsart}
\usepackage{geometry}                
\usepackage{graphicx}
\usepackage{amssymb}
\usepackage{epstopdf}
\usepackage[superscript,biblabel]{cite}
\title{Artificial Intelligence in Humans}
\author{Dr. Michael Swan Laufer Ph.D.}

\begin{document}
\maketitle
\section{Introduction}
In the early days of artificial intelligence, Alan Turing put forward a test in hopes of answering the question as to whether machines can think\cite{turing}. It continues to this day to be the benchmark test for artificial intelligences; now called the Turing test. The test consists of running a series of conversational exchanges over a text channel between a human tester and other humans, as well as the AI in question. If the human tester cannot reliably identify which of the conversations are the AI, the AI has passed the Turing test.

Fifty years later, Edward Feigenbaum, put forward a variation of this test, in which a sophisticated, specialized AI can be tested against a team of specialists in the particular area in which the AI has been trained\cite{feigenbaum}.

In Feigenbaum's paper, he describes the Turing test as follows: \\

\begin{quote}
\emph{In the test, a human judgement must be made concerning whether a set of observed behaviors is sufficiently similar to human behaviors that the same word---intelligent---can justifiably be used. The judgement is about behaviors not mechanism. Computers are not like human brains, but if they perform the same acts and one performer (the human) is labeled intelligent, then the other must be labeled intelligent also.} \\
\end{quote}  
The interesting result which this reasoning sires is: since this judgement is about behaviors and not mechanisms, the only thing which truly separates a human from a machine is the mechanism by which we think. The important point here being that the machine imitates behaviour without ``understanding" what is occurring. 

This issue has long been a hot point of debate in the philosophy of AI. The ``strong AI" theory asserts that an AI is capable of having a mind, as well as mental states; versus the ``weak AI" position, which suggests that a machine is only capable of behaving intelligently.

Strangely, there is a parallel in the world of pedagogy, which seems to remain unexplored. When we work to create curricula and motivate methods of instruction, far too often there is a focus on behavior, rather than the mechanisms within a student which motivate their behavior.

However, students, teachers, and schools are all measured based primarily on standardized test scores\cite{gould}.  This is understandable, because, ideally, it allows for an unbiased, quantifiable, scale with which one can compare achievement of teachers and ability of students. The shortcoming, however, is that as soon as these tests are made the measure of worth, then everyone involved shifts their focus from mastering the material upon which the test is based, to mastering the test itself. This objection has been well-explored\cite{ayers}\textsuperscript{,}\cite{popham}\textsuperscript{,}\cite{davidson}\,.

If this were the only issue, solutions could be crafted in order to fix this; but there is a deeper problem here. All too often students are learning an abstract set of behaviors which imitate understanding of the material. Additionally, teachers are training this very thing actively, because it is a safe strategy. Attempts to foster understanding may or may not yield results; training however always does (provided students are sufficiently hardworking, obedient, and have suitable medium-term memories.) Furthermore, in some instances, the teachers themselves lack understanding of the material, but are merely adept at training these imitative behaviours.

If we teach students to mimic behaviours in order to make them perform well on assessments, we rob them of their humanity. Worse than that: we waste it. We reduce people to a state of nonthinking, in which they have to work extremely hard to stick to protocols which may be extremely complex, and we do not provide any means to understand the ideas upon which those protocols are based.

Feigenbaum goes on to say:\\
\begin{quote}
\emph{For ... a computational intelligence, to be able to behave with high levels of perfomance on complex intellectual tasks, ... it must have extensive knowledge of the domain.} \\
\end {quote} 

 This is essentially code for what every struggling teacher already knows: if your students do not understand the concepts you are trying to teach them, you must get them to memorize a tremendous amount of information in order to compensate. This is the root of the tendency for to approach to all subjects to become content-based. 

\section{Continuation}
When he goes on to define ``Subject Matter Expert Turing Test", which we now call the Feigenbaum Test, he asks if a human judge can determine, at better than chance level, which entity is his colleague, and which is the computer, after asking deep probing questions about a specific area of specialized expertise. If one replaces the computer in this scenario with a human impostor, we are faced with the situation	 which I put forward in my paper \emph{A Misanthropic Interpretation of the Chinese Room Problem}\cite{laufer}. If a person has merely been trained to give responses to queries on a particular topic the way an expert would, it might be quite difficult to determine that the person was not, in fact, a genuine expert. If we extend this idea to pedagogy, the question arises as to whether or not a human judge (much less an exam) can differentiate between a student who has a genuine understanding of the material, and one who merely has filled their memory with so many work-around mechanisms that they can imitate the behaviour of a student who is understanding.

The obvious answer is no. Or at least, not without great difficulty. If we were to implement assessments to differentiate between these two, I fear we would be extremely disappointed, and would find that the number of students who are actually understanding, rather than imitating well is negligible. 


We are essentially programming people as though they were high powered AI computers, or what Feigenbaum terms \emph{expert systems}; a programmed protocol for obtaining outcomes based on inputs in an efficient and predictable manner. We then declare ourselves successful, when we give them what is essentially a simplified Feigenbaum test, and those whom we have trained perform well. This is almost tautological. We are merely training people to perform well on exams. The instances when someone performs well on an exam because they posses an understanding of the material, is almost a fluke. 

We cheat all students this way. The ones who do not wish to learn, we reward when they bypass the learning process with exam mechanisms. To the ones who genuinely understand the material, we do not give sufficient credit, because their performance on exams does not differ from those who merely use test taking strategies and memory. Worst of all, those students who are struggling to understand the material, and have a genuine interest will oftentimes fall back to strategies of exam performance because of the pressure caused by evaluation based on examination. Rare indeed it is when a student would sacrifice their grades for the possibility of understanding.

As tragic as all of this is, it is entirely understandable. Both sides of the interchange have the best of intentions: the students want to perform well, and the teachers want to have a large quotient of well-performing students. Furthermore, the technique of training a student for an exam is a stable, robust solution, and so it is attractive. It has consistent and medium to fair results in the aforementioned measured outcomes. 

\section{Why this is important.}
We need to shift our focus from evaluating the bahaviours of students to evaluating the mechanisms by which they produce their outcomes. If one takes this idea to a slightly less abstract level, one can imagine asking a person with no understanding of arithmetic to learn how to perform two-digit multiplication; they leave and come back after some time, and give the correct answers to every given question every time. This may seem satisfactory. However, if we then ask this person how they are performing the task, and they say merely that they memorized all 10,000 possible outcomes of two digit multiplication, you would be very disappointed that they didn't really learn what you were trying to convey. 
I ask, though: when a student is taught a procedure for two-digit multiplication, for which the basis in the distributive property is not explained, is this really much better, save for the fact that there is less tax on their memory? Don't get me wrong: mechanisms and shortcuts are useful for making the mechanics of many things faster and easier. However if mechanisms are not grounded in concept, we are merely programming artificial intelligence modules into human brains. The human brain is capable of so much more, and it is truly a crime that we use such a powerful and elegant tool in such a crude way.

I wish to stress that this is not an epistemological question in the way some of these questions can be when they are posed in the AI setting; we are dealing with real human beings, who can explain their thinking, and their methods of arriving at their answers. 

Of course, were we dealing with a theoretical savant who had a perfect-recall memory, and spent years reading a dozen books a day on a given topic, and was able to merely recall the appropriate passage from some text each time they were prompted in casual conversation, it would be difficult, and perhaps take quite a long time before they finally tripped up, and said something that alerted you to their underlying ignorance. However in most cases, if we were set out with the intention of determining whether someone had genuine understanding of a topic or not, it undoubtedly would not take long at all to uncover their level of mastery, merely by going conceptually deeper with each question.

\section{Solutions}
In order to fix this problem we must work hard to base any content on concept in the process of conveying it. In the course of examination, in addition to asking a student to complete a task, we have to ask in an abstract narrative form for the student to explain why they took the approach they did, and why it works. This solution is far from popular, because it involves a tremendous amount of work. 
However if we are genuinely trying to help students learn, and not merely trying to perpetuate a tradition of faith in what was said by those who came before, we have an obligation to put in any amount of work which is required in order to help.

\section{Conclusions}
Unless we stop treating our students like machines, and start addressing the mechanisms of their thought processes, we may as well stop teaching altogether, and apply all our efforts to developing AI systems, which will run in just as sterile a manner, but faster, and with greater reliability. 

We must focus on those things which set us apart from machines, and develop those actively, or else we would do better to work on replacing ourselves with machines entirely, and bring an end to the era of humans to a close.

\vspace{\fill}

\newpage
\clearpage

\end{document}